\documentclass[11pt,journal,epsfig,onecolumn,peerreviewca,draftclsnofoot]{IEEEtran}
\usepackage{amsmath}
\usepackage{amssymb}
 \usepackage{stfloats}
 \usepackage{makecell}

\usepackage{graphicx}
\usepackage{epstopdf}
\usepackage{subfigure}
\usepackage{epstopdf}
\usepackage{todonotes}
\usepackage{algorithm}
\usepackage{algorithmic}
\usepackage{framed}
\usepackage{comment}
\usepackage{subeqnarray}
\usepackage{stfloats}
\usepackage[justification=centering]{caption}
\usepackage{cite}
\ifCLASSOPTIONcompsoc
 \usepackage[caption=false,font=normalsize,labelfont=sf,textfont=sf]{subfig}
\else
 \usepackage[caption=false,font=footnotesize]{subfig}
\fi

\begin{document}
\title{Supervised and Semi-Supervised Deep Neural Networks for CSI-Based Authentication}
\author{Qian~Wang,~Hang~Li, Zhi~Chen,~\IEEEmembership{Senior Member,~IEEE}, Dou~Zhao, Shuang~Ye, and Jiansheng~Cai
\thanks{The authors are with the National Key Laboratory of Science and Technology on Communications, University of Electronic Science and Technology of China, Chengdu 611731, China (e-mail: wangqianfresh@hotmail.com, hanglee94@hotmail.com, chenzhi@uestc.edu.cn, ZhaoDou\_UESTC@outlook.com, yesh\_425@outlook.com, jason199502@outlook.com).}}
\maketitle
\begin{abstract}
From the viewpoint of \emph{physical-layer authentication}, spoofing attacks can be foiled by checking channel state information (CSI). Existing CSI-based authentication algorithms mostly require a deep knowledge of the channel to deliver decent performance. In this paper, we investigate CSI-based authenticators that can spare the effort to predetermine channel properties by utilizing deep neural networks (DNNs). We first propose a convolutional neural network (CNN)-enabled authenticator that is able to extract the local features in CSI. Next, we employ the recurrent neural network (RNN) to capture the dependencies between different frequencies in CSI. In addition, we propose to use the convolutional recurrent neural network (CRNN)---a combination of the CNN and the RNN---to learn local and contextual information in CSI for user authentication. To effectively train these DNNs, one needs a large amount of labeled channel records. However, it is often expensive to label large channel observations in the presence of a spoofer. In view of this, we further study a case in which only a small part of the the channel observations are labeled. To handle it, we extend these DNNs-enabled approaches into semi-supervised ones. This extension is based on a semi-supervised learning technique that employs both the labeled and unlabeled data to train a DNN. To be specific, our semi-supervised method begins by generating pseudo labels for the unlabeled channel samples through implementing the K-means algorithm in a semi-supervised manner. Subsequently, both the labeled and pseudo labeled data are exploited to pre-train a DNN, which is then fine-tuned based on the labeled channel records. Finally, simulation and experimental results show that the proposed DNNs-enabled schemes can significantly outperform the benchmark designs, and their semi-supervised extensions can yield excellent performance even with limited labeled samples.
\end{abstract}

\section{Introduction}
\subsection{Physical Layer Authentication}
With the advent of the 5th generation network, an enormous amount of private and confidential information, e.g., financial data, medical records, and customer files, will be transmitted via the wireless medium \cite{Yang2015Safeguarding}. The sharp increase in demand for wireless security continuously requests more advanced authentication schemes. Traditionally, authentication mechanisms are performed above the physical layer by using secret keys to identify wireless transmitters. Despite their effectiveness, they are faced with two main challenges: On the one hand, the high key management overhead results in concerns such as excessive latencies. On the other hand, the time required to crack a key has been remarkably shortened with the growing processing power; see a recent overview \cite{WangMag} and the references therein. 
The idea of \emph{physical-layer authentication} is to validate a wireless transmitter by verifying the physical-layer attributes of the wireless transmission. In comparison to conventional secret key-based authentication schemes, \emph{physical-layer authentication} needs no key distribution and management. Besides, it is extremely difficult to impersonate a wireless transmission's physical-layer features. Thanks to these facts, \emph{physical-layer authentication} is deemed as a promising technique to make the unrivalled security service a reality. Some of the existing \emph{physical-layer authentication} approaches rely on the analog front-end imperfections, which are \emph{device-specific} characteristics caused by manufacturing variability \cite{MerchantRFdeeplearning}. \emph{Device-specific} characteristics, such as in-phase/quadrature imbalance \cite{HaoIQimbalance}, the power amplifier characteristics \cite{PolakTransImperfect}, and the carrier frequency offset \cite{HouCFO}, have relatively stable nature. However, the difference of the targeted hardware features between devices is usually too small in practice, which will be further influenced by noise and interference \cite{WangMag}. Another class of physical-layer features used for authentication purposes are channel-based characteristics, like channel state information (CSI) \cite{XiaoTimeVariant,XiaoFrequencyselective,Liu2D} and received signal strength (RSS) \cite{YchenWSN}. CSI is hard to predict due to the presence of rich scatters and reflectors in a general wireless communication environment. Besides, it is safe to say that users located at different places have uncorrelated channels. These facts make CSI a \emph{location-specific} characteristic that has aroused great interest for user authentication.


Paper \cite{XiaoTimeVariant} studied CSI-based authentication in a time-variant wireless environment, wherein the channel variation was modeled with the assumption of a first-order autoregressive model. The authors in \cite{XiaoFrequencyselective} studied frequency-selective channels, in which the terminal mobility-caused channel variation was modeled as a first-order autoregressive model, and the environment changes and the estimation errors were modeled as independent complex Gaussian processes. Moreover, a two-dimensional (the dimensions of channel amplitude and path delay) quantization method was proposed in \cite{Liu2D} to preprocess the channel variations, wherein the temporal processes were still modeled as autoregressive models. To sum up, existing CSI-based approaches \cite{XiaoTimeVariant,XiaoFrequencyselective,Liu2D} formulated the authentication process as binary hypothesis testing by exploiting the correlation between CSI at adjacent times. All these works designed algorithms that would look for predetermined features in CSI. To do this, the system operator needs to possess sufficient channel information such as the channel model and the channel variation pattern. This kind of authentication system will be vulnerable to small ambiguities in the a priori messages.


Recently, machine learning techniques have found their applications in the realm of \emph{physical-layer authentication}. Particularly, authors in \cite{XiaoReinforcementLearning} investigated the RSS-based authentication game in a dynamic environment, in which \emph{reinforcement learning} was utilized to achieve the optimal test threshold in the hypothesis test. Paper \cite{MerchantRFdeeplearning} used time-domain complex baseband error signals to train a convolutional neural network (CNN) so that user identities can be derived based upon \emph{device-specific} imperfections. The logistic regression model was utilized in \cite{XiaoLandmarks} to exploit the received signal strength indicators measured at multiple landmarks for user identification. Although the spatial resolution of the transmitter can be enhanced through using multiple landmarks, their deployment will raise the system overhead and more pressingly, the communication between the landmarks and the security agent will be confronted with severe security threats. Thankfully, CSI contains much more \emph{location-specific} information than RSS and can thus be reliable enough without assistances such as landmarks.
\subsection{Our work}
In this paper, we establish deep neural networks (DNNs)-enabled authenticators that connect a transmitter's CSI to its estimated identity. DNNs have been extensively studied and found to be very effective in learning high-level features from raw data for objects identification \cite{RichFeature,AKimagenet}. The CNN was first proposed for digit recognition \cite{Yann1998} and later became one of the most widely applied DNNs. It usually utilizes multiple convolutional layers that can successively generate deeper-level abstractions of the input data. The key of implementing CSI-based authentication lies in the correlation between the channel observations for the same user at different times. However, this correlation can be weakened by factors such as environment changes and practical imperfections. Fortunately, the CNN can be invariant to the transformations of channel observations resulting from these factors; hence we propose to exploit the CNN to extract the deep features in CSI for user authentication. Also, we try to analyze CSI from a sequential point of view. In this way, CSI is seen as a data sequence and we utilize a recurrent neural network (RNN) to model the dependencies between different frequencies in CSI. The RNN is mainly designed for sequence modeling \cite{Graves2013Speech}. It employs feedback loops to allow connections from previous states to the subsequent ones and thus is able to represent advanced patterns of dependencies in the sequence. As a matter of fact, the CNN and the RNN possess different modeling abilities. More concretely, the CNN is good at representing locally invariant information while the RNN is better at contextual information modeling. Based on this observation, we propose to use the convolutional recurrent neural network (CRNN) \cite{Shiendtoend,WuCRNN} for CSI-based authentication. The CRNN is an emerging deep model that contains both the CNN and the RNN so that it can exploit not only the representation power of the CNN but also the contextual information modeling ability of the RNN. Accordingly, it is expected that the CRNN can have advantages over the CNN and the RNN in modeling channel features for authentication, which is confirmed by the simulation and experimental results given in section \ref{SimRes}.

To efficiently train the proposed DNNs-enabled authenticators, a large amount of labeled channel observations are required. Although it is easy to obtain adequate channel observations since channel estimation is an indispensable part of wireless communication, it is expensive to label large observations with the adversary being very crafty. This motivates us to consider the \emph{physical-layer authentication} further in a severe circumstance where the number of labeled channel observations is limited. In order to deal with the small sample problem, we propose to use semi-supervised learning. The original idea of semi-supervised learning, known as \emph{self-training}, can be traced back to 1960s, which appeared as a wrapper-algorithm that repeatedly employs a supervised learning technique \cite{Chapelle}. Semi-supervised learning algorithms have been widely investigated these years \cite{Ranzato,Rasmus,LeePseudoLabel,WuFuzzy}. Employing DNNs in semi-supervised learning is an emerging trend \cite{Ranzato,Rasmus}. A DNN-enabled authenticator can be trained in a semi-supervised manner with both the labeled and unlabeled data. This can be done by assigning pseudo labels to the unlabeled samples, which are then exploited as if they are actually labeled \cite{LeePseudoLabel,WuFuzzy,ss-Kmeans}. The pre-trained network can be further fine-tuned by only using the labeled channel observations \cite{ZuoCCRN}.

Specifically, this work considers a wireless system in which the service agent aims to validate the access right of a wireless transmitter via examining its CSI. It should be mentioned that there is no need to make assumptions on the channel model or the channel variation since our interest lies on data driven self-adaptive algorithms that demand no a priori knowledge of the underlying channel properties. Our main contribution includes the following aspects:
\begin{enumerate}
\item To begin with, we build a CNN-enabled classifier. The main components of this classifier are convolutional layers, which are able to generate deep-level \emph{feature maps} through locally convolving small subregions of its input. Also, the network employs pooling layers to subsample the output of the convolutional layers so as to reduce the computational complexity and avoid over-fitting. At the end of the network, we use fully connected layers, with one logistic layer on the top, to collect the early extracted features and learn the user identity.
\item Next, we establish a RNN-enabled authenticator that analyzes CSI from a sequential perspective. This authenticator is composed of several recurrent layers and a few fully connected layers. The recurrent layers use feedback loops to capture the spectral dependencies in CSI, which are then input to the following fully connected layers such that object recognition can be implemented. 
\item We then propose a CRNN-enabled approach that works in the following way: The first part of the proposed authenticator is a CNN, which is used for extracting middle-level features. Next, the output features of the CNN are fed into recurrent layers so that the contextual information of CSI can be well captured. Finally, fully connected layers are employed to perform classification.
\item 
    Furthermore, we extend the proposed DNNs-enabled authentication methods to deal with a small sample problem, i.e., only a small number of channel observations are labeled. This extension is based on a semi-supervised scheme that takes advantage of limited labeled channel samples and abundant unlabeled channel observations simultaneously. In our semi-supervised approach, the K-means algorithm is first employed to assign pseudo labels to the unlabeled data. With the existence of some labeled samples, the K-means algorithm is implemented in a semi-supervised manner. Subsequently, we propose to utilize both the labeled and pseudo labeled data to pre-train a DNN and then perform fine-tuning using only the labeled channel observations.
\item Lastly, we use simulations and experiments to demonstrate the performance of the proposed methods. Both the simulation and experimental results show that the proposed DNNs-enabled algorithms can achieve significant performance gains over the benchmark schemes, and the CRNN-enabled authenticator owns the highest accuracy and the shortest convergence time simultaneously among the proposed DNNs-enabled approaches. Also, it is presented in the simulation results that the proposed semi-supervised DNNs-enabled approaches can achieve excellent authentication accuracy even when the number of labeled channel observations is limited.
\end{enumerate}



\section{System Model and Problem Statement}
\begin{figure}[h]
\centering
\includegraphics[width=9cm]{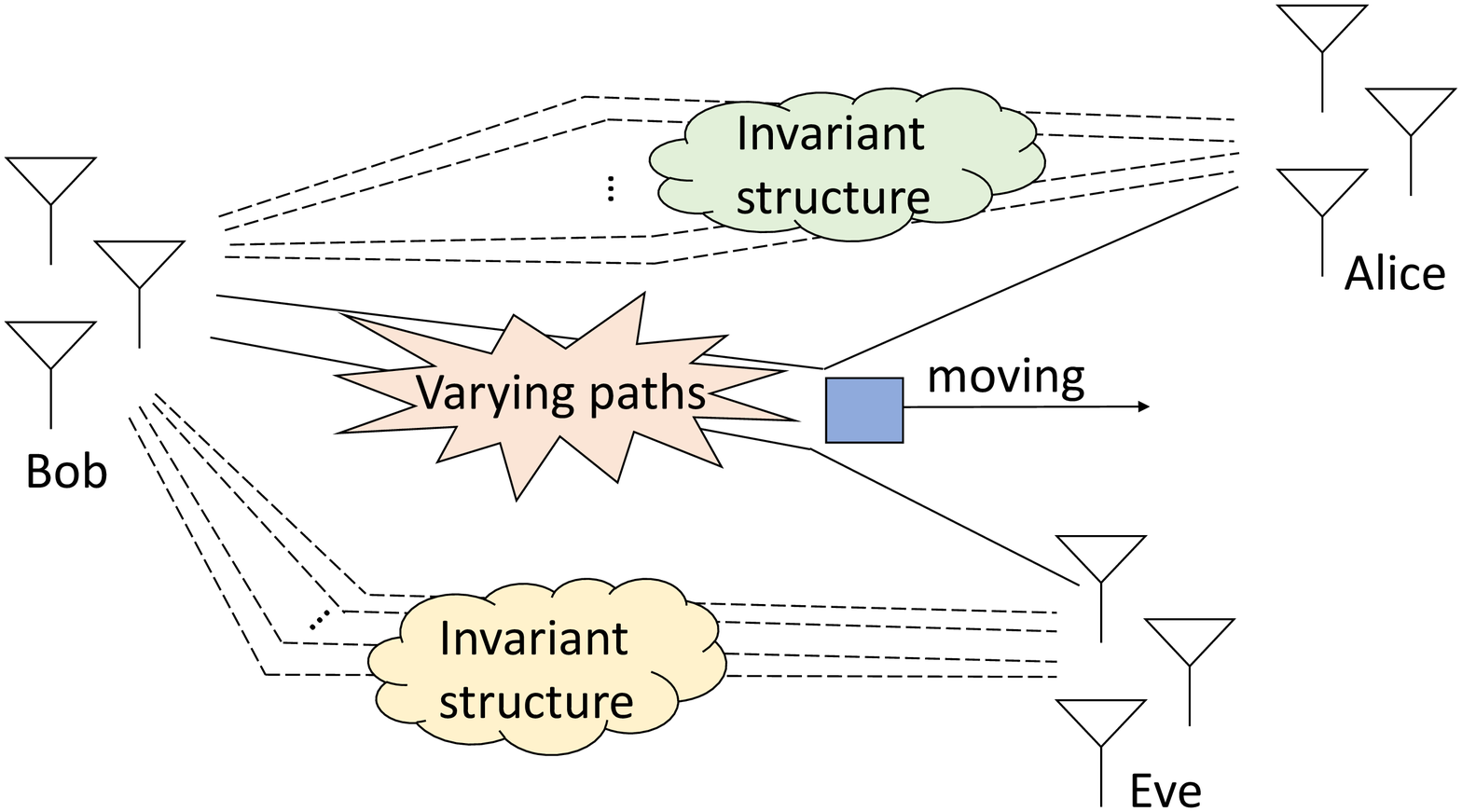}
\DeclareGraphicsExtensions.
\caption{Illustration of our considered three-node communication system.} \label{systemmodel}
\end{figure}

Consider a typical ``Alice-Bob-Eve" network shown in Fig. \ref{systemmodel}, in which Bob is entasked with the job of providing services for both Alice and Eve, while Eve is unauthorized as far as the secure service intended for Alice is concerned. We assume that Bob is equipped with $M_{B}$ antennas, both Alice and Eve have $M_{A}$ antennas, and CSI is measured at $N$ tones. In our setup, Alice, Bob, and Eve are geographically placed at different locations. Also, suppose that Alice, Bob, and Eve stay stationary in a time-variant communication environment. This is common in practical scenarios where one may put one's cellphone/laptap on the phone stand/desk while using it, and the service agent is stationary by nature. As an untrusted user, Eve may impersonate Alice by forging the digital credential of Alice, like the password, the IP address, and the MAC address, in attempts to illegitimately acquire confidential information intended for Alice or send false messages to Bob. Once Eve successfully obtains the illegal advantages, the following attacks will be devastating so that the authentication process is of paramount importance for the network security.

Bob aims to foil the spoofing attacks launched by Eve through establishing a physical-layer authentication mechanism. Specifically, Bob intends to verify whether the transmitter who uses Alice's digital credential is Alice or not by carefully checking its CSI according to the historical CSI of Alice and Eve. To do this, it is supposed that Bob can record historical CSI and the corresponding user identities. This is possible because both Alice and Eve are network users. Generally, device mobility and environment changes are the factors that give rise to the channel variation. Since all the communication nodes are assumed to be static in this work, it is safe to say that environment changes are the only reasons that bring about the channel variation. Environment changes such as the movement of objects can affect part of the existing paths while other paths stay invariant. Take an indoor environment for example, there may be moving people and objects, but the ceiling, the floor, walls, and furniture will always stay still. This fact indicates that the channel between two static terminals has an invariant structure, which forms the foundation of our application of deep learning into the CSI-based authentication problem.

This work is focused on CSI-based authentication that maps a transmitter's CSI to its authenticated identity, i.e.,
\begin{align}\label{fa}
\hat{I}_t=f_a(\textbf{H}(t)),
\end{align}
in which $\textbf{H}(t)\in\mathcal{C}^{M_B\times M_A\times N}$ denotes the communication channel observed at time $t$,  $\hat{I}_t$ represents the corresponding estimated identity, and $f_a$ is the authenticator that can distinguish the legitimate channel from the illegitimate channel. In our settings, $\hat{I}_t=1$ indicates that the estimated transmitter at time $t$ is Eve, and $\hat{I}_t=0$ means that Alice is the estimated transmitter at time $t$. Additionally, $I_t=1$ denotes that the transmitter at time $t$ is Eve, and $I_t=0$ represents that Alice is the transmitter at time $t$. The aim in this paper is to build authenticators that can accurately determine the authenticity of a transmitter without the demand to predetermine the underlying channel properties. With no a priori information about the channel observations, we propose to model $f_a$ with neural networks, which are efficient models for statistical pattern recognition. The structure of $f_a$ can be settled when the network architecture is specified. Through training the network parameters with historical CSI, $f_a$ can be fully derived.

\section{Deep Neural Networks for CSI-Based Authentication}
In this section, we introduce DNNs to construct CSI-based authenticators that can capture the invariant channel structure from noisy historical CSI instead of checking prearranged characteristics. Specifically, the introduced DNNs are the CNN in Section \ref{subsectionCNN}, the RNN in Section \ref{subsectionRNN}, and the CRNN in Section \ref{subsectionCRNN}. These DNNs count upon different mechanisms to model the invariant channel features, while all of them employ fully connected layers to perform classification based on the extracted information. A fully connected layer can be expressed as
\begin{align}
\boldsymbol{v}=f_v(\textbf{W}_{gv}\boldsymbol{g}),
\end{align}
in which $\boldsymbol{g}$ denotes the early captured feature sequence, $\textbf{W}_{gv}$ is a transformation matrix, $f_v$ represents an activation function, and $\boldsymbol{v}$ can be hidden units or the predicted class label. The final layers of a DNN are fully connected layers whose activation functions are chosen as rectified linear units (ReLUs), with one logistic layer on the top producing the authentication result.

For ease of the subsequent description, we define $f_d(\textbf{H}(t),\boldsymbol{w})$ as the network function of a DNN, where all the weights and biases are grouped together into a vector $\boldsymbol{w}$. Since the activation function of the output layer is a logistic sigmoid, we have $0\le f_d(\textbf{H}(t),\boldsymbol{w})\le1$. One can interpret $f_d(\textbf{H}(t),\boldsymbol{w})$ as the conditional probability $p(I_t=1|\textbf{H}(t),\boldsymbol{w})$, with $p(I_t=0|\textbf{H}(t))$ derived as $1-f_d(\textbf{H}(t), \boldsymbol{w})$. The conditional distribution of the class label given the input channel is a Bernoulli distribution, i.e.,
\begin{align}
p(I_t|\textbf{H}(t),\boldsymbol{w})=f_d(\textbf{H}(t),\boldsymbol{w})^{I_t}[1-f_d(\textbf{H}(t),\boldsymbol{w})]^{1-{I_t}}.
\end{align}
 Accordingly, a DNN-enabled authenticator can be written as
\begin{align}
 f_a(\textbf{H}(t))=\left\lceil{f_d(\textbf{H}(t),\boldsymbol{w})-1/2}\right\rceil,
\end{align}
in which $\left\lceil{x}\right\rceil$ denotes the ceiling function that maps $x$ to the least integer greater than or equal to $x$. 

Given a training set of channels $\{\textbf{H}(t)\}_{t=1}^{T}$, together with a corresponding set of labels $\{I_t\}_{t=1}^{T}$, in which $T$ is the number of training samples, we train the network to minimize the error function, which is a \emph{cross-entropy} error function of the form
\begin{align}\label{Ew}
E(\boldsymbol{w})=-\sum_{t=1}^{T}\{I_t\ln f_d(\textbf{H}(t),\boldsymbol{w})+(1-I_t)\ln[1-f_d(\textbf{H}(t),\boldsymbol{w})]\}.
\end{align}
We use the stochastic gradient descent (SGD) method to train the network. The gradients in the convolutional layers and the recurrent layers are calculated by the backpropagation algorithm and the backpropagation through time (BPTT) algorithm, respectively. To reduce the error fluctuation, our implementation utilizes a mini-batch strategy, that is, the gradients are calculated based on mini-batches. $\boldsymbol{w}$ will be iteratively updated until the training and validation loss converges.

\subsection{CNN}\label{subsectionCNN}
\begin{figure}
  \centering
  \includegraphics[width=9cm]{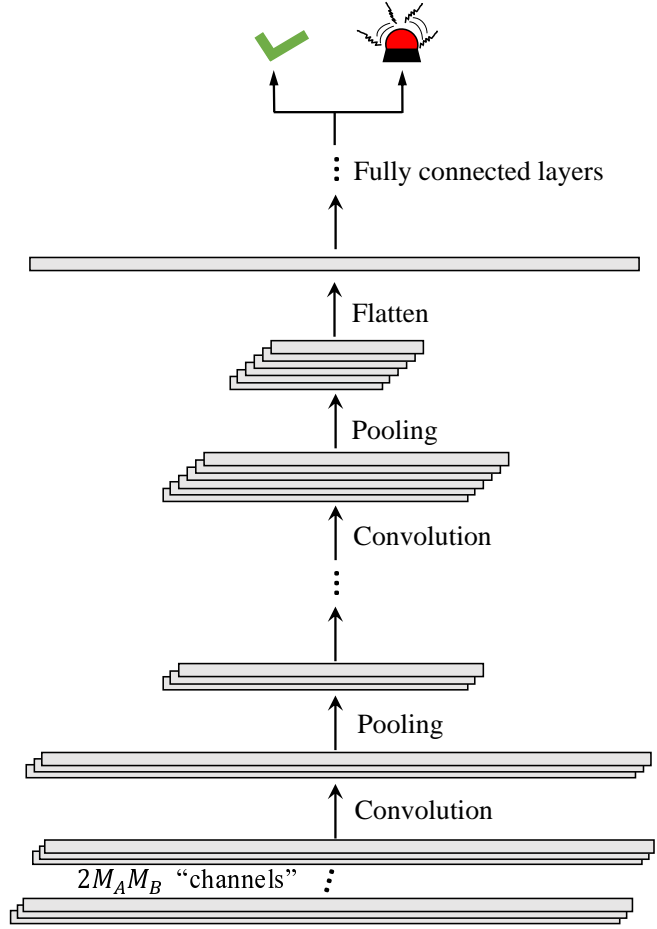}
   \DeclareGraphicsExtensions. \caption{Illustration of our employed architecture of the CNN.}\label{figCNN}
\end{figure}
As illustrated above, the time-variant channel has an invariant structure. In this subsection, we employ the CNN to extract the invariant channel features from varying channel observations, based on which the fully connected layers can implement the classification operation. In Fig. \ref{figCNN} we plot the architecture of our employed CNN, which is consisted of convolutional layers, pooling layers, and fully connected layers. Being the core components of the CNN, convolutional layers utilize the following mechanisms:
\begin{itemize}
\item[$\bullet$] Local Receptive Fields - The input of the convolutional layer is divided into local receptive fields (small subregions), each of which is connected to a single neuron of the next layer. As a benefit of this, the number of connections, as well as the number of parameters, is drastically cut down in the convolutional layer.
\item[$\bullet$] Weight Sharing - Each locally applied connection is essentially a filter. A convolutional layer employs multiple filters, which are reused over all the local receptive fields. The reuse of filters leads to the sharing of an identical set of weights among different connections.
\end{itemize}
Invoking the above mechanisms, a convolutional layer organizes its input units into \emph{feature maps}, i.e.,
\begin{align}
\boldsymbol{F}=(\boldsymbol{f}_1,\boldsymbol{f}_2,...,\boldsymbol{f}_N)=f_f(\boldsymbol{s}*\{\boldsymbol{\phi_1},\boldsymbol{\phi_2},...,\boldsymbol{\phi_d}\}),\label{g}
\end{align}
where $\boldsymbol{f}_n\in\mathcal{R}^d, n=1,2,...,N,$ represents a \emph{feature map}, $f_f$ is an activation function, $\boldsymbol{s}$ is the input vector, $*$ denotes a convolution, and $\{\boldsymbol{\phi_1},\boldsymbol{\phi_2},...,\boldsymbol{\phi_d}\}$ is a set of filters.
Units in a \emph{feature map} take input from a local receptive field of $\boldsymbol{s}$, and different receptive fields share the same filters.

Notice that before feeding the network input into the first convolutional layer, we organize it into $2M_AM_B$ ``channels'',\footnote{When a neural network is utilized to analyze an image, there will be three input ``channels'' corresponding to the red, green, and blue elements of the input image, respectively.} each of which corresponds to the real or imaginary part of the channel between a pair of transmitting and receiving antennas. To be specific, the set of all the filters is repeatedly applied to all these ``channels'' and the results for different ``channels'' are added before input to the activation function $f_f$. There are numerous nonlinear activation functions applicable to the neural network framework, such as the hyperbolic tangent function, the softmax, and the ReLU, in which the ReLU is the most widely applied in the CNN and thus is chosen to be the activation function $f_f$ in (\ref{g}).

We perform pooling to summarize the \emph{feature maps} created in the convolutional layer. Specifically, each pooling unit takes input from a region in the corresponding \emph{feature map}, which is called a \emph{pooling window}. The commonly used pooling operations are \emph{max pooling}, \emph{average pooling}, and \emph{stochastic pooling}. By utilizing pooling, the neural network can achieve more compact representations that are more robust to noise and interference. In our architecture, the size of the \emph{pooling window} is set as $1\times3$, on which a $\max$ operation is implemented. One can see that the features extracted by the convolutional layers and the pooling layers contain multiple ``channels'', which are flattened before fed into the fully connected layers, i.e., multiple ``channels'' are used in series.
\subsection{RNN}\label{subsectionRNN}
 \begin{figure}[h]
\centering
\includegraphics[width=9cm]{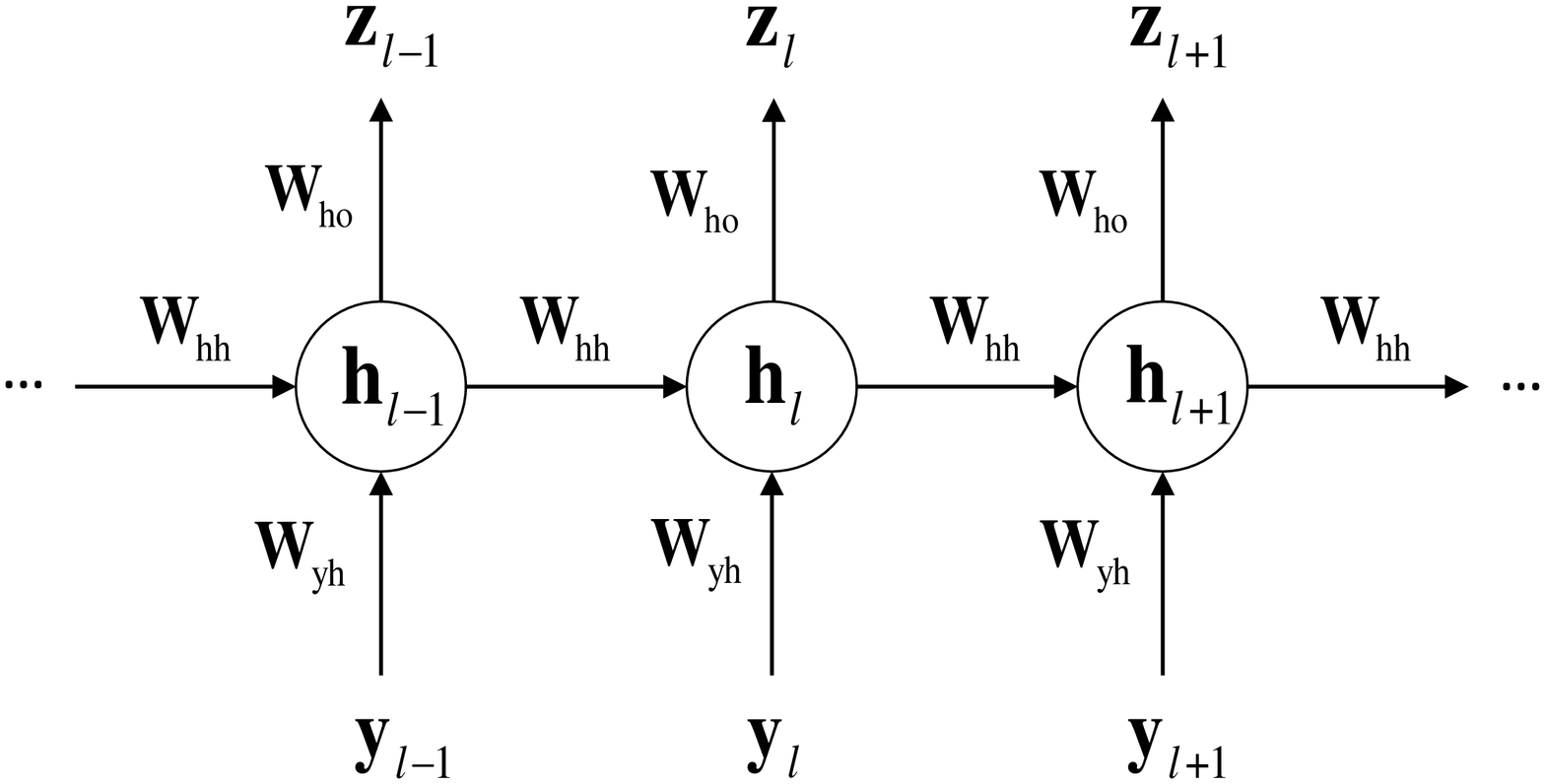}
\DeclareGraphicsExtensions.
\caption{The structure of a recurrent layer.} \label{figRNN}
\end{figure}
Due to the fact that there exist spectral dependencies in the channel, our goal in this subsection is to analyze the channel from a sequential point of view. Towards this end, we use the RNN to capture the contextual information in the channel for the purpose of authentication. The RNN utilizes feedback loops in its recurrent layers to connect the previous states with the current ones. A graphical illustration of a recurrent layer is shown in Fig. \ref{figRNN}. When a sequence is processed with length $L$, its hidden feature $\boldsymbol{h}_l$ and the predicted output $\boldsymbol{z}_l$ at stage $l\in[1,...,L]$ are derived as
\begin{subequations}
\begin{align}
&\boldsymbol{h}_l=f_h(\boldsymbol{W}_{hh}\boldsymbol{h}_{l-1}+\boldsymbol{W}_{yh}\boldsymbol{y}_l),\label{hl}\\
&\boldsymbol{z}_l=f_z(\boldsymbol{W}_{hz}\boldsymbol{h}_{l})\label{zl},
\end{align}
\end{subequations}
respectively, in which $\boldsymbol{y}_l$ denotes the $l$th input, $\boldsymbol{W}_{hh}$, $\boldsymbol{W}_{yh}$, and $\boldsymbol{W}_{hz}$ are transformation matrices, and $f_h$ and $f_z$ are activation functions. With the existence of feedback loops, a recurrent layer is able to memorize the historical information so that they can discover meaningful connections between a single data and its context. In our architecture, the recursive function $f_h$  and the activation function $f_z$ are chosen to be the hyperbolic tangent function and the logistic sigmoid, respectively. It should be pointed out that as a $M_B\times M_A\times N$ complex channel, the network input is flattened before it is fed into the first layer of the RNN.

The RNN we use in this work first employs several recurrent layers to capture spectral dependencies in its input. Then, fully connected layers are utilized to implement classification based on the early extracted features that contain the contextual information.
\begin{figure}
  \centering
  \includegraphics[width=9cm]{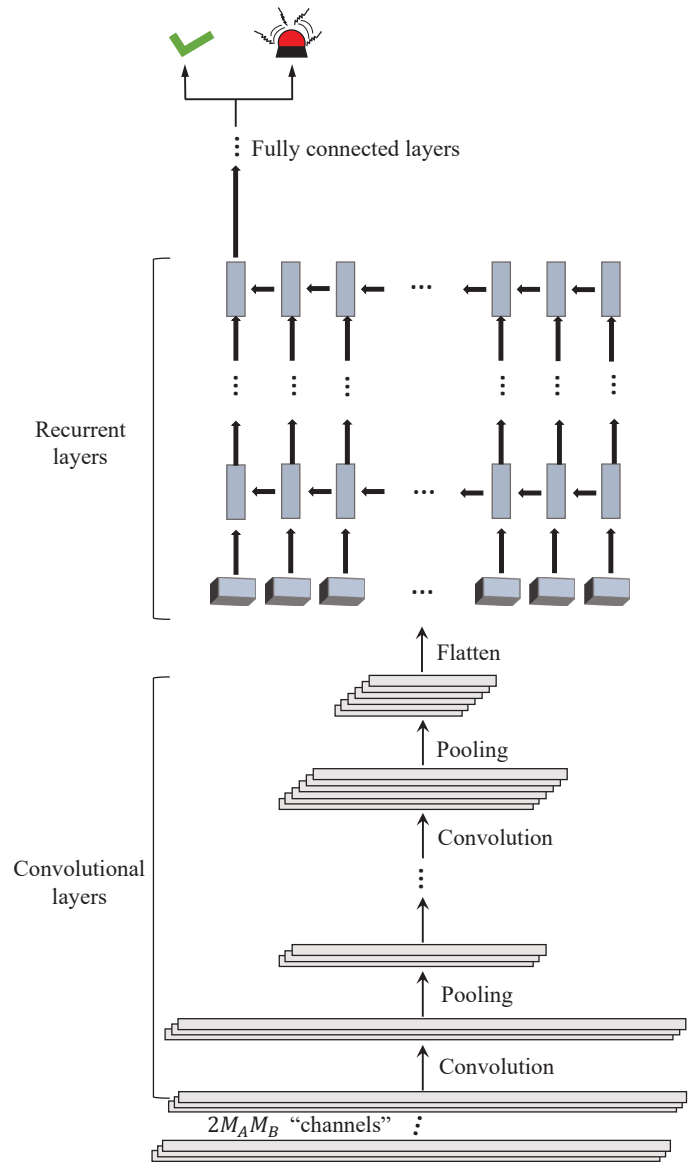}
   \DeclareGraphicsExtensions. \caption{Illustration of our employed architecture of the CRNN.}\label{CRNN}
\end{figure}
\subsection{CRNN}\label{subsectionCRNN}
So far, we have introduced the CNN and the RNN for CSI-based authentication. It is easy to notice that these two DNNs have distinct modeling abilities since they rely on different mechanisms. To be specific, the CNN is good at capturing locally invariant features while the RNN is adept at contextual information extraction. In this subsection, we propose to utilize a hybrid of the CNN and the RNN, i.e., the CRNN, that combines the abilities of the CNN and the RNN such that the deep features containing both the locally invariant information of CSI and the contextual messages between different frequencies in CSI can be well extracted and further be exploited for user authentication.

 As schematically illustrated in Fig. \ref{CRNN}, our employed CRNN is consisted of multiple convolutional layers (together with pooling layers), several recurrent layers, and a few fully connected layers. The mechanisms of these layers have been discussed above. In the CRNN, the convolutional layers can capture middle-level features, which are useful for the dependencies modeling at the recurrent layers. At the same time, the contextual information learned by the recurrent layers can lead to better representations at the convolutional layers during backpropogation. The CRNN takes good advantage of both the discriminative representation capability of the CNN and the contextual information extraction power of the RNN and is therefore expected to outperform both of them in CSI-based authentication.

\section{Extensions of the Proposed DNNs-Enabled Authenticators for a Small Sample Problem}
In this section, our focus lies on extending the proposed DNNs-enabled authentication methods into semi-supervised approaches to deal with a small sample problem. To efficiently train a DNN, a large amount of labeled samples are required. Since channel estimation is necessary for wireless communication, it is not difficult to acquire adequate channel observations. However, the labeled channel records may be limited because the labeling work could be quite expensive in the presence of an attacker who can play tricks. In view of this, we propose a semi-supervised learning technique to deal with the case where there are abundant channel observations but only a small part of them are labeled. The key idea of semi-supervised learning is to make use of both the labeled and unlabeled samples so that the data set is large enough to effectively train a DNN. To describe the proposed semi-supervised technique, we denote the channel samples available at Bob as $\{\Omega_U,\Omega_L,\Psi_L\}$, in which $\Omega_U$ and $\{\Omega_L,\Psi_L\}$ are unlabeled and labeled channel records, respectively.
 \begin{figure}[h]
\centering
\includegraphics[width=9cm]{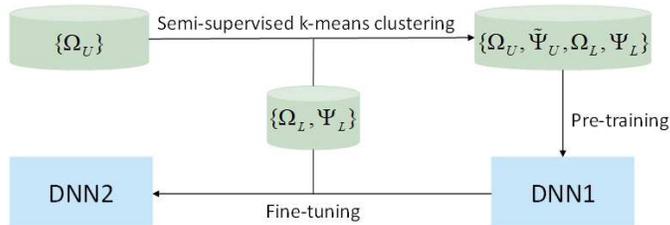}
\DeclareGraphicsExtensions.
\caption{The framework of the proposed semi-supervised learning method.} \label{SSDL}
\end{figure}

The framework of the proposed semi-supervised learning method is illustrated in Fig. \ref{SSDL}. As one can see from the figure, the network begins by generating pseudo labels $\tilde{\Psi}_U$ for the unlabeled channel records $\Omega_U$ through using a semi-supervised k-means algorithm. Next, the labeled and unlabeled channel records $\{\Omega_U,\Omega_L,\Psi_L\}$, together with the pseudo labels $\tilde{\Psi}_U$, are used to train a DNN. This pre-training process leads to an immature DNN, which is denoted as DNN1. Lastly, the labeled samples $\{\Omega_L,\Psi_L\}$ are exploited to fine-tune DNN1 so as to obtain a fully developed DNN, i.e., DNN2. In the sequel, we will elaborate on the generation of pseudo labels and the pseudo labels-aided semi-supervised deep learning.
\subsection{Generation of Pseudo Labels}\label{ss_Kmeans}
It is obvious that unlabeled data cannot be used to train the neural network directly. In order to exploit the unlabeled channel samples in the network training process, we assign pseudo labels to them by employing clustering. The channel observations at Bob should be categorized as two clusters, each of which corresponds to a particular transmitter, i.e., Alice or Eve. To be specific, we utilize k-means clustering to separate the unlabeled channel records into two groups. When implementing clustering, the labeled channel records are utilized to provide class information so that the unlabeled channel observations can be classified into labeled categories. The performance of the k-means algorithm is dominated by the selection of initial centres. Generally, the k-means algorithm is initialized with randomly chosen centres, which may incur the error floor effects. As far as our scenario is considered, redundant centre updates can be avoided in the clustering task due to the existence of some labeled data \cite{ss-Kmeans}. In our proposed semi-supervised k-means algorithm, the centre of a cluster is initialized with the centroid of the corresponding labeled channel records. The detailed procedure of the proposed semi-supervised k-means method is summarized in Algorithm \ref{algorithm1}.
\begin{algorithm}
\renewcommand{\algorithmicrequire}{\textbf{Input:}}
\renewcommand\algorithmicensure {\textbf{Output:} }
\caption{The proposed semi-supervised k-means algorithm}
\begin{algorithmic}[1]
\STATE \textbf{Input}: $\{\Omega_U,\Omega_L,\Psi_L\}$.
\STATE Let $\boldsymbol{c}_j, j\in\{0,1\}$ be the centroid of the channel records labeled with $j$.
\STATE Let the set of centers be $\mathcal{C}=\{\boldsymbol{c}_0,\boldsymbol{c}_1\}$.
\STATE \textbf{repeat}
\STATE \quad Assign each $\boldsymbol{H}_i\in\Omega_U, i=1,2,..., card(\Omega_U)$ a pseudo label $\tilde{I}_i$ according to the nearest center $\boldsymbol{c}\in\mathcal{C}$.
\STATE \quad Update $\boldsymbol{c}_j, j\in\{0,1\}$ as the centroid of records with label $j$ or pseudo label $j$.
\STATE \quad Update $\mathcal{C}=\{\boldsymbol{c}_0,\boldsymbol{c}_1\}$.
\STATE\textbf{until} $\mathcal{C}$ not changed.
\STATE \textbf{Output}: $\tilde{\Psi}_U=\{\tilde{I}_i\}_{i=1}^{card(\Omega_U)}$.
\end{algorithmic}
\label{algorithm1}
\end{algorithm}
\subsection{Semi-Supervised Deep Learning Using Pseudo Labels}
\begin{figure}[h]
\centering
\includegraphics[width=10cm]{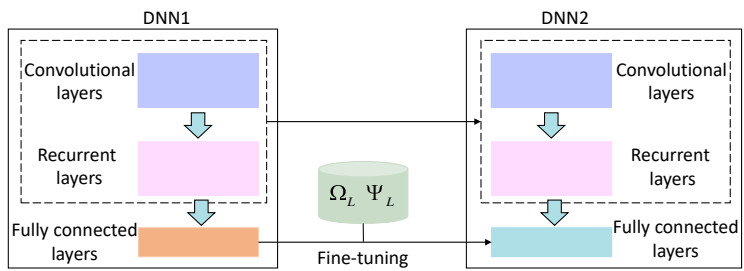}
\DeclareGraphicsExtensions.
\caption{The fine-tuning stage for the semi-supervised framework.} \label{fine_tune}
\end{figure}

As illustrated in Section \ref{ss_Kmeans}, we have obtained the pseudo labels $\tilde{\Psi}_U$ for the unlabeled observations $\Omega_U$. Now, we can use both the labeled and pseudo labeled data, i.e., $\{\Omega_U,\tilde{\Psi}_U,\Omega_L,\Psi_L\}$, to train a DNN such that the training data set is large enough. Although a pseudo labeled record has a certain degree of fuzziness in its class information, the pseudo labels mostly coincide with the unknown true labels. As a benefit of this, the deep features can still be preserved and can thus be extracted by the hidden layers of DNN1. Then we use the labeled records $\{\Omega_L,\Psi_L\}$ to fine-tune DNN1. It is clear that the CNN and the RNN can be deemed as special cases of the CRNN. Thus we shall only discuss the fine-tuning process of the CRNN. In our implementation, the process of fine-tuning is directly conducted on DNN1, during which the parameters in the convolutional layers and recurrent layers stay frozen and only the parameters of the fully connected layers are updated due to the fact that we only have a small amount of labeled samples. The fine-tuning stage for the semi-supervised framework is presented in Fig. \ref{fine_tune}. As plotted, the fully developed DNN, i.e., DNN2, has the same network structure with DNN1, and their convolutional layers and recurrent layers are identical.
\section{Simulation and Experimental Results}\label{SimRes}
In this section, we first generate a simulation dataset based upon practical assumptions and then use Monte Carlo simulations to demonstrate the performance of the proposed algorithms on the simulation dataset and compare them with some benchmark designs. Next, experiments based on Universal Software Radio Peripherals (USRPs) \cite{SM} are conducted to validate the efficacy of the proposed methods on real data, in which the benchmark designs are tested again for comparison.
\subsection{A Simulation Dataset}\label{dataset}
In this subsection, a simulation dataset is created according to practical assumptions. Consider a multipath fading channel model with an exponentially decaying power-delay profile, i.e., the average channel power at delay $\tau$ can be written as $\frac{1}{\sigma_\tau}e^{-\tau/\sigma_\tau},$ where $\sigma_\tau$ denotes the root mean square delay spread. The total number of paths is calculated as $p_{max}=\left\lceil {\frac{10\sigma_\tau}{T_s}}\right\rceil$, in which $T_s$ represents the sampling period and is set to be $150$ns. Suppose that the power of the $p$th channel tap is a complex Gaussian random variable with zero mean and variance $\sigma^{2}_{p}=\sigma^2_{0}e^{-pT_s/\sigma_\tau},$ wherein
$\sigma^2_{0}=\frac{1-e^{-T_s/\sigma_\tau}}{1-e^{-(p_{max}+1)T_s/\sigma_\tau}}.$ All the terminals are assumed to be placed in a typical office and $\sigma_\tau$ is accordingly set as $50$ns \cite{EldadPerahia2008Next}, which is shared by the legitimate and illegitimate channels.  

The settings for the simulation dataset are as follows: The number of antennas at Bob is $M_B=3$. Alice and Eve both have $M_A=1$ antenna. CSI is measured at $N=128$ tones. For simplicity, the distance between adjacent antenna elements is assumed to be larger than a half of the wavelength so that the channels between different transmitting and receiving antenna pairs are independent with each other. Denote $\check{\boldsymbol{h}}_{A,k}(t)\in\mathcal{C}^{N\times1}$ and $\check{\boldsymbol{h}}_{E,k}(t)\in\mathcal{C}^{N\times1}, k=1,2,3,$ as the channels from the $k$th antenna at Bob to Alice and to Eve, respectively, at time $t$. This work considers a dynamic environment. Analogous to \cite{XiaoFrequencyselective}, we model $\check{\boldsymbol{h}}_{i,k}(t), i\in\{A,E\}, \forall k, t,$ as the sum of an average gain $\bar{\boldsymbol{h}}_{i,k}$ and a random variation $\boldsymbol{\epsilon}_{i,k}(t)\sim \mathcal{CN}(\boldsymbol{0},{\delta}_1^2\boldsymbol{\text{I}}_N)$, in which $\mathcal{CN}(\boldsymbol{0},\boldsymbol{\text{C}})$ represents the complex Gaussian distribution with zero mean and covariance matrix $\boldsymbol{\text{C}}$, $\boldsymbol{\text{I}}_n$ denotes the identity matrix of order $n$, and $\boldsymbol{\epsilon}_{i,k}(t)$ is assumed to be independent across different $\{i,k,t\}$. In particular, the assumption of the complex Gaussian distribution can be justified by the following facts. On one hand, the channel variation is a complex variable since the channel is complex. On the other hand, various unpredictable sources of changes make the path variation a random variable. For a large number of variations on independent paths, the sum of these variations can be modeled as a Gaussian variable based on the Central Limit Theorem.

 Given the fact that CSI can not be perfectly estimated since there are always undesirable factors such as interference, thermal noise, and hardware imperfections, we model the channel observation $\tilde{\boldsymbol{h}}_{i,k}(t), \forall i, k, t,$ as the sum of $\check{\boldsymbol{h}}_{i,k}(t)$ and the channel estimation error $\boldsymbol{\varepsilon}_{i,k}(t)\sim \mathcal{CN}(\boldsymbol{0},{\delta}_2^2\boldsymbol{\text{I}})$, which is assumed to be independent across different $\{i,k,t\}$. For notational convenience, we define the sum of $\boldsymbol{\epsilon}_{i,k}(t)$ and $\boldsymbol{\varepsilon}_{i,k}(t)$ as the observed variation $\boldsymbol{\xi}_{i,k}(t), \forall i, k, t$, the covariance matrix of which is set as $50\boldsymbol{\text{I}}$. To perform Monte Carlo simulations, we consider numerous channel trials in the simulation dataset. For each channel trial, the average gain $\bar{\boldsymbol{h}}_{i,k}, \forall i,k,$ is randomly generated. 
\subsection{Simulation Setup}
\begin{table}
\centering
\caption{Summary of the Network Configurations}
\begin{tabular}{c|c|c|c|c|c} 
\hline
CRNN-4&CNN-4&RNN-4&CRNN-3&CNN-3&RNN-3\\
\hline
conv1x3-32&conv1x3-32&recur-256&conv1x3-32&conv1x3-32&recur-128\\
maxpooling&conv1x3-32&recur-256&maxpooling&conv1x3-32&recur-256\\
conv1x3-64&maxpooling&recur-512&conv1x3-64&maxpooling&recur-512\\
maxpooling&conv1x3-64&recur-512&maxpooling&conv1x3-64&FC-512-64\\
recur-256&conv1x3-64&FC-512-64&recur-512&maxpooling&FC-64-1\\
recur-512&maxpooling&FC-64-1&FC-512-64&FC-2048-64&\\
FC-512-64&FC-2048-64&&FC-64-1&FC-64-1&\\
FC-64-1&FC-64-1&&&&\\
\hline
\end{tabular}\label{NetworkSettings}
\end{table}
In our setup, each DNN has two fully connected layers, and each CNN and CRNN employs two pooling layers. Specifically, CRNN-3 has two convolutional layers and one recurrent layers, CNN-3 has three convolutional layers, RNN-3 has three recurrent layers, CRNN-4 has two convolutional layers and two recurrent layers, CNN-4 has four convolutional layers, and RNN-4 has four recurrent layers. The configurations of these networks are summarized in Table \ref{NetworkSettings}, in which ``conv$1\times3$-$n_1$'' denotes a convolutional layer with a receptive field size of $1\times3$ and $n_1$ filters, ``maxpooling'' is a maxpooling layer, ``recur-$n_2$'' represents a recurrent layer whose feature dimension is $n_2$, and ``FC-$n_3$-$n_4$'' denotes a fully connected layer with $n_3$ input units and $n_4$ output units. We employ a workstation with two 1.7-GHz Intel(R) Xeon(R) E5-2603 v4 CPUs to perform simulations, in which the neural networks are implemented based on the TensorFlow framework \cite{TensorFlow}. During each training process, $10\%$ of the training samples are utilized to validate the hyperparameters such as the learning rate and the mini-batch size. The test set utilized to examine the performance contains $200$ labeled channel samples per class.
\subsection{Performance for the DNNs-Enabled Authentication Algorithms on the Simulation Dataset}
In this subsection, Monte Carlo simulations are employed with $100$ channel trials to demonstrate the performance of the proposed DNNs-enabled authentication methods on the simulation dataset. For each trial, we randomly select $500$ labeled channel records per class to form the training set.  At the beginning of the training process, the network weights are randomly chosen, and the learning rate is set to be $10^{-4}$. Then, the mini-batch SGD algorithm is run for $100$ epoches to update the network parameters, in which the batch size is set to be $256$ and the learning rate halves every $20$ epoches.
\begin{figure}
  \centering
  \subfigure{
  \begin{minipage}{9cm}
  \centering
  \includegraphics[width=9cm]{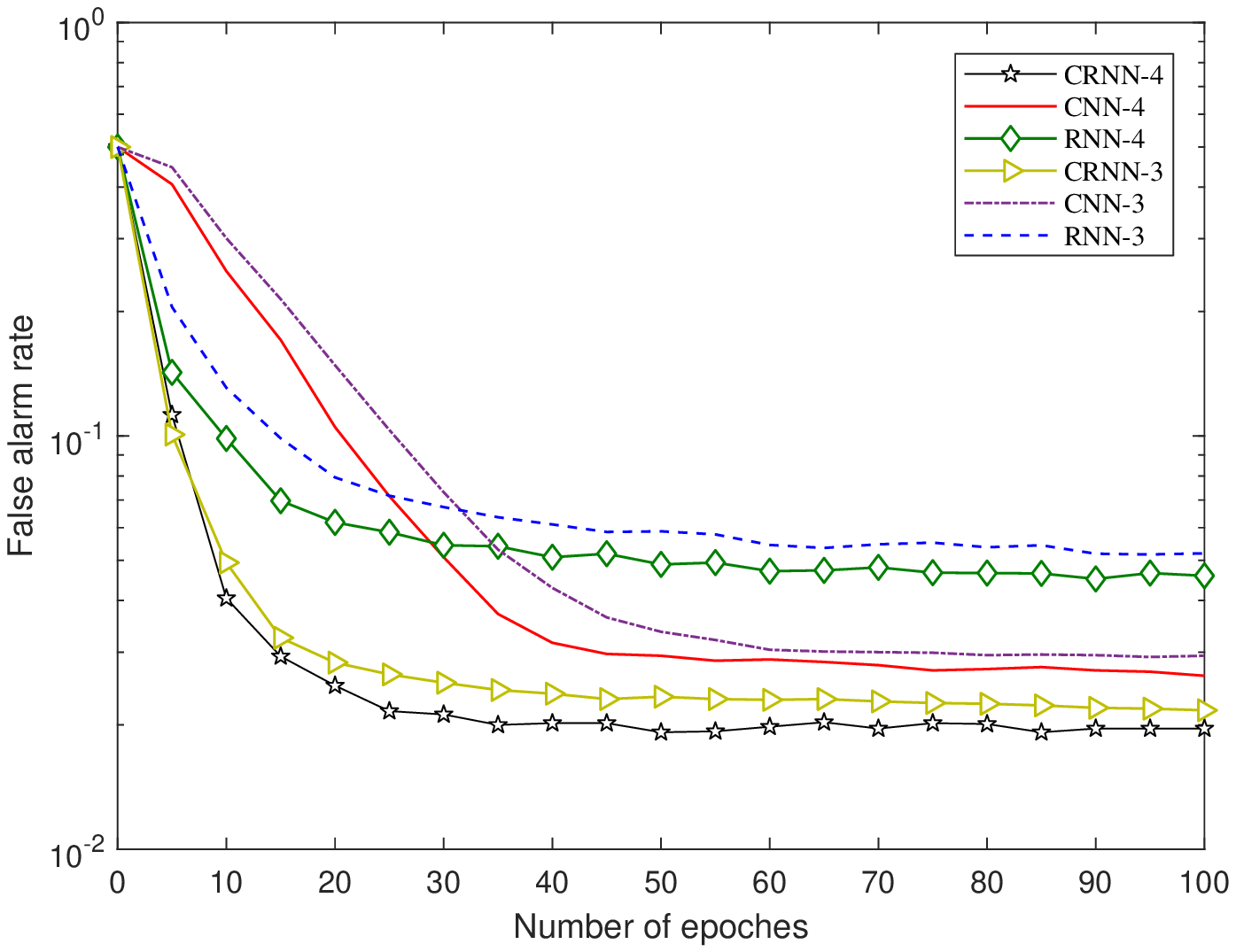}
  \end{minipage}
  }

  \subfigure{
  \begin{minipage}{9cm}
  \centering
  \includegraphics[width=9cm]{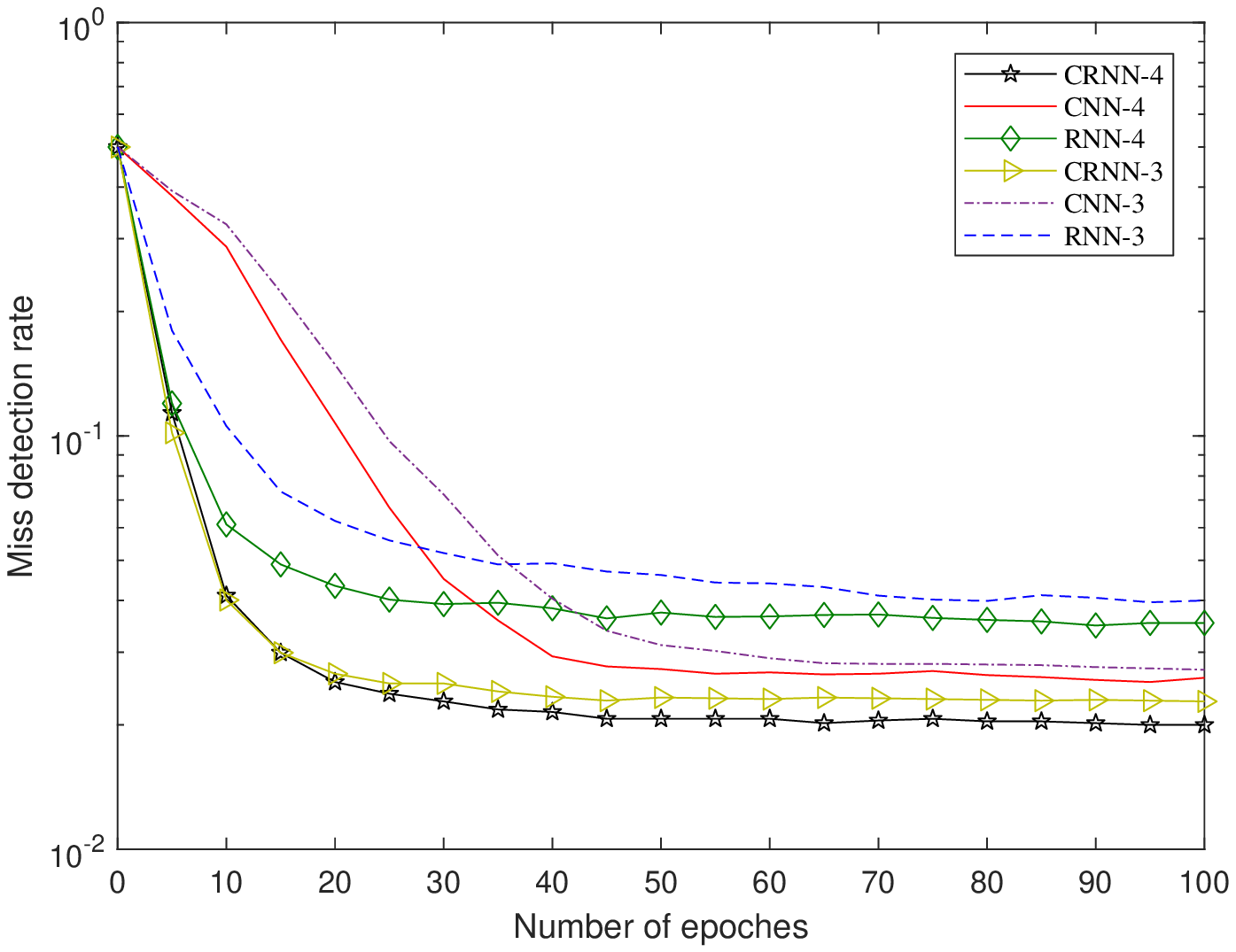}
  \end{minipage}
  }
  \DeclareGraphicsExtensions. \caption{False alarm rates and miss detection rates for different DNNs.}\label{falsemiss}
  \end{figure}

In Fig. \ref{falsemiss} we give the false alarm rates and the miss detection rates achieved by different DNNs with respect to (w.r.t.) the number of epoches. From the figure, one can see that as the number of epoches grows, every DNN has a converged false alarm rate and a converged miss detection rate. It is presented in the figure that for the same type of DNN, the deeper model is more beneficial to the authentication performance. This is because the deeper model can be trained to extract more discriminative features than the shallower one. As seen from the figure, although the CNN can achieve a lower false alarm rate and a lower miss detection rate than the RNN, the CNN with additional recurrent layers, i.e., the CRNN, can significantly outperform that with additional convolutional layers. This is expected since the CRNN is armed with the ability of the RNN in addition to that of the CNN, while the CNN can only capture local features no matter how deep the network is.
 \begin{figure}[h]
\centering
\includegraphics[width=9cm]{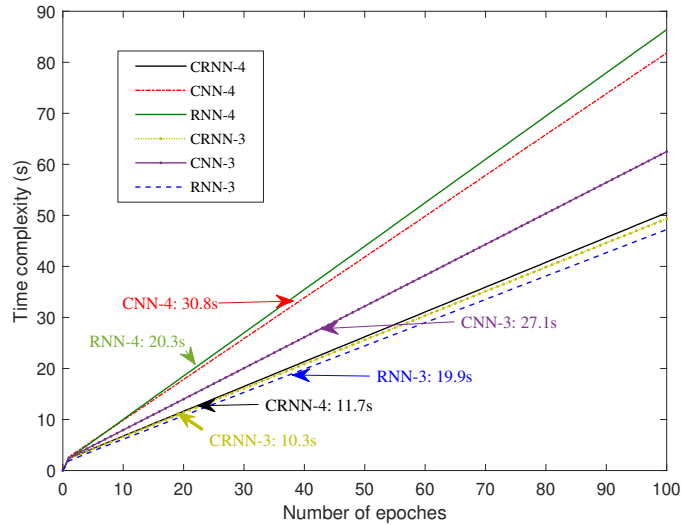}
\DeclareGraphicsExtensions.
\caption{Time complexity for different DNNs.} \label{time}
\end{figure}

Fig. \ref{time} shows the time complexity for different DNNs w.r.t. the number of epoches, wherein we use several arrows to pinpoint their convergence times, respectively. One can see from the figure that for the same class of DNN, the deeper model converges slower than the shallower one, which can be deemed as the cost required to get the performance gain over the latter. Also, it is viewed that the CNN leads to better authentication performance than the RNN at the expense of increased computation overhead. Moreover, one can notice that the CRNN converges faster than both the CNN and the RNN. This observation, together with the simulation results plotted in Fig. \ref{time}, illustrates that the combination of the CNN and the RNN can not only improve the authentication performance but also accelerate the convergence, which makes the CRNN a success in CSI-based authentication.
 \begin{figure}[h]
\centering
\includegraphics[width=9cm]{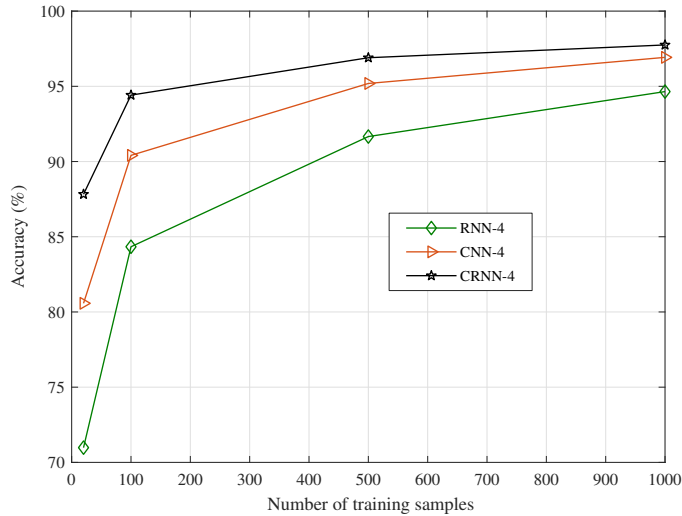}
\DeclareGraphicsExtensions.
\caption{The authentication accuracy for different DNNs-enabled algorithms.} \label{diffsamples}
\end{figure}

We plot in Fig. \ref{diffsamples} the authentication accuracy for different DNNs-enabled algorithms w.r.t. the number of training samples. As observed, the authentication accuracy increases with the number of training samples for all the DNNs-enabled methods. This is due to the fact that with a lager number of training samples, a DNN can extract more discriminative information for user authentication. It is also seen that regardless of the sample number, the CRNN-enabled approach always yields the highest authentication accuracy among these DNNs-enabled algorithms. Moreover, one can notice that the CRNN-enabled approach is the most robust one whose performance decline is the smallest when the number of training samples decreases, while the RNN-enabled algorithm is the most vulnerable one who has the biggest drop in accuracy as training samples are reduced. With a smaller number of training samples, different DNNs-enabled algorithms have larger performance gaps.

Table \ref{table1} presents the authentication results for different methods. Apart from the proposed DNNs-enabled approaches, we also consider the usage of the logistic sigmoid, the K-nearest neighbor (KNN) algorithm \cite{Zhang2007ML}, the support vector machine (SVM) \cite{BSVM}, and a NP test for comparison. Since we have already established DNNs-enabled authenticators, it is not difficult to see how to build CSI-based authenticators by exploiting the logistic sigmoid, the KNN, and the SVM. The detailed process is omitted due to the page limit. The NP test we consider is given by
\begin{align}\label{NP}
L\triangleq||\boldsymbol{H}-(\bar{\boldsymbol{h}}_{A,1},\bar{\boldsymbol{h}}_{A,2},\bar{\boldsymbol{h}}_{A,3})||^2\gtrless\gamma,
\end{align}
 wherein $||\boldsymbol{M}||$ is the Frobenius norm of the matrix $\boldsymbol{M}$, $\boldsymbol{H}\in\mathcal{C}^{N\times M_B}$ denotes the to-be-authenticated channel observation, $\gamma$ represents a specially chosen threshold, and  $\bar{\boldsymbol{h}}_{A,i}, \forall i,$ is assumed to be known. The estimated identity will be Eve if $L>\gamma$, otherwise the transmitter will be authenticated as Alice. It needs to be mentioned that the authentication accuracy varies with $\gamma$ and the results given in Table \ref{table1} are obtained with the optimal $\gamma$.
\begin{table}
\centering
\caption{Authentication Results for Different Methods}
\begin{tabular}{c|c|c} 
\hline
Method&Accuracy (\%)&Time Complexity (s)\\
\hline
NP test&64.9&$3\times10^{-6}$\\
\hline
KNN&90.1&9.9\\
\hline
Logistic&89.9&30.6\\
\hline
SVM&93&127.5\\
\hline
RNN-4&94.7&20.3\\
\hline
CNN-4&96.9&30.8\\
\hline
CRNN-4&97.8&11.7\\
\hline
\end{tabular}\label{table1}
\end{table}

As shown in the table, there are huge performance gaps between the NP test and the machine learning-based methods. This is because one may need a priori channel information to design a NP test-based algorithm that can deliver decent performance, while the machine learning-based algorithms can analyze the invariant channel structure intelligently from historical CSI. It is obvious that the time complexity of the intelligent algorithms far exceeds that of the NP test. Luckily, the network training process can be done ahead of time and thus the immediate authentication response is possible. In Table \ref{table1}, it is seen that all the DNNs-enabled algorithms can achieve significantly higher authentication accuracy than the benchmark approaches. This is attributed to the fact that our adopted DNNs not only have remarkable modeling abilities but also apply to the CSI-based authentication problem. 
\subsection{Performance for the Semi-Supervised DNNs-Enabled Authentication Methods on the Simulation Dataset}
This subsection studies the performance of the proposed semi-supervised DNNs-enabled methods on the simulation dataset. The achieved accuracy is calculated based on an average of $100$ Monte Carlo runs as before. Since the semi-supervised DNNs-enabled methods aim to deal with the small sample problem, the labeled set only has $10$ channel observations per class in each channel trail. To do semi-supervised learning, we also utilize a large amount of unlabeled data, the size of which is set to be $1000$. This semi-supervised scheme begins by assigning pseudo labels to the unlabeled channel records through invoking Algorithm \ref{algorithm1}. Next, the labeled and the pseudo labeled channel samples are combined to per-train a DNN. During this pre-training process, the network weights are randomly initialized. Then, we run the mini-batch SGD algorithm for $100$ epoches and the batch size is set to be $256$. The learning rate is initialized as $10^{-4}$, which halves every $20$ epoches. After the pre-training process is done, we can derive DNN1, which is then fine-tuned to obtain DNN2. During the fine-tuning process, the parameters in the convolutional layers and the recurrent layers are excluded from updating, and we use the labeled samples to train the fully connected layers of DNN1. The fine-tuning process runs the SGD algorithm for $100$ epoches and the initial learning rate is set as $10^{-3}$ to speed up the training, which decays by half every $20$ epoches.
\begin{table}
\centering
\caption{Comparison of Supervised and Semi-Supervised Methods}
\begin{tabular}{c|c|c} 
\hline
Method&Accuracy (\%)&Time Complexity (s)\\
\hline
CNN-4 (semi)&96.3&45.6\\
\hline
CNN-4&80.6&10.7\\
\hline
RNN-4 (semi)&93.3&37.5\\
\hline
RNN-4&71&9\\
\hline
CRNN-4(semi)&97.1&30.2\\
\hline
CRNN-4&87.8&5.5\\
 \hline
\end{tabular}\label{table2}
\end{table}

In Table \ref{table2} we give the authentication results for the supervised and semi-supervised DNNs-enabled algorithms. As expected, the effectiveness of these (supervised) DNNs-enabled methods are greatly reduced when the number of labeled training samples is limited. This is because that DNNs can not be efficiently trained when training samples are insufficient. By comparing Table \ref{table1} and  Table \ref{table2}, it is noticed that the authentication accuracy of a semi-supervised DNN-enabled method with $20$ labeled samples and $1000$ unlabeled samples is lower than that achieved by its supervised counterpart with $1000$ labeled samples. This performance gap results from the fuzziness existed in the pseudo labels. Despite the fuzziness, one can see from Table \ref{table2} that a semi-supervised DNN-enabled algorithm can perform significantly better than its supervised counterpart when they are handling the small sample problem. This observation leads to a conclusion that our proposed semi-supervised approach can significantly amplify the abilities of these DNNs-enabled authentication algorithms through utilizing limited labeled channel samples and abundant unlabeled channel samples at the same time. Besides, Table \ref{table2} presents that in both the supervised and semi-supervised cases, the CRNN-enabled method can not only deliver the best authentication performance but also own the shortest convergence time compared to the CNN-enabled algorithm and the RNN-enabled approach.
\subsection{Experimental Results}
 \begin{figure}[h]
\centering
\includegraphics[width=9cm]{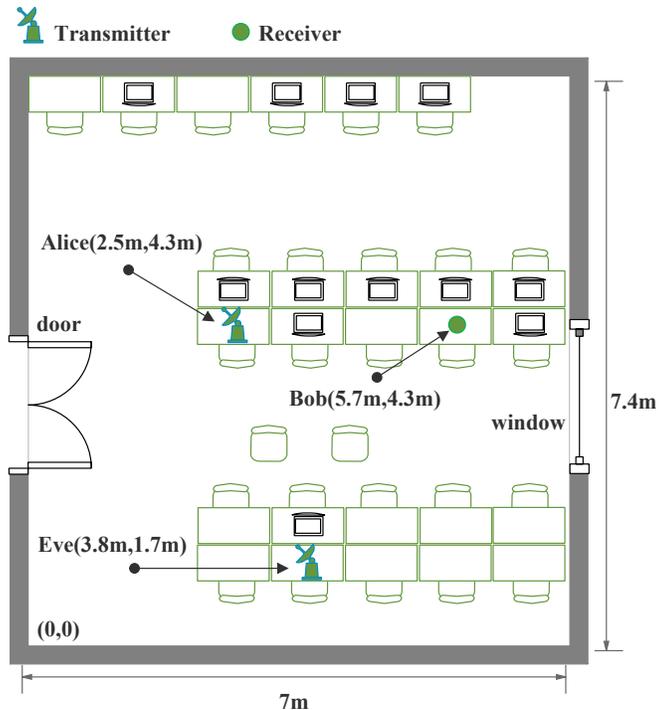}
\DeclareGraphicsExtensions.
\caption{The experimental scenario.} \label{experiment}
\end{figure}
As illustrated in Fig. \ref{experiment}, our experiments are conducted in a $7.4\times7\times5\text{m}^3$ office room, where there are two transmitters, i.e., Alice and Eve, and one receiver, i.e., Bob. Each transmitter is $3.2$m away from the receiver. Specifically, we utilize a USRP-2955, a USRP-2954R, and a USRP-2944R to work as Bob, Alice, and Eve, respectively. Every USRP is connected to a computer through a Peripheral Component Interconnect Express bus. We exploit Laboratory Virtual Instrumentation Engineering Workbench \cite{LabVIEW} to generate the digital baseband transmission signal. The USRP-2954R and the USRP-2944R are used to perform digital-to-analog conversion and frequency upconversion on the digital baseband signal and then send it via an omnidirectional antenna. After receiving the radio frequency signal, the USRP-2955 downconverts it to baseband and does analog-to-digital conversion on it. The experimental settings are as follows: Each USRP employs a single antenna. The transmit power is equal to $0.6$mW for both the USRP-2954R and the USRP-2944R. The communication bandwidth is $1$MHz with the carrier frequency at 5GHz. Both the Alice-to-Bob channel and the Eve-to-Bob channel are estimated by the least squares method at $900$ different times. The channel observations obtained at Bob are separated into a training set and a test set, wherein the training set contains $500$ channel observations per class, while the test set has $400$ channel observations per class.
\begin{table}
\centering
\caption{Summary of the Network Configurations for Real Data}
\begin{tabular}{c|c|c} 
\hline
CRNN-r&CNN-r&RNN-r\\
\hline
conv1x3-32&conv1x3-32&recur-128\\
maxpooling&maxpooling&recur-256\\
recur-128&FC-2048-64&FC-256-64\\
FC-128-64&FC-64-1&FC-64-1\\
FC-64-1&&\\
\hline
\end{tabular}\label{NetworkSettingsr}
\end{table}

Since the hyperparameters of a neural network should be validated according to the training samples and the channel observations acquired in the experiments are different from that in the simulation dataset, the network hyperparameters generated based on the simulation dataset may not apply here. Therefore, we redetermine the network configurations and summarize them in Table \ref{NetworkSettingsr}. During the network training process, the network parameters are randomly initialized and the initial learning rate is set as $10^{-3}$. Next, the network parameters are updated by running the mini-batch SGD algorithms for $100$ epochs with a batch size of $256$, wherein the learning rate halves every $20$ epoches.

We show in Table \ref{tableRealData} the authentication results for different methods on real data. Again, the KNN algorithm, the Logistic sigmoid, and the SVM are introduced as benchmarks. As one can see from the table, all the proposed DNNs-enabled authenticators can perform significantly better than the benchmark designs. Besides, the CRNN-enabled method can achieve the highest authentication accuracy and the shortest convergence time at the same time among the propose DNNs-enabled algorithms. These results agree with that obtained on the simulation dataset. By comparing the authentication performance given in Table \ref{table1} and Table \ref{tableRealData}, one can get an interesting observation that the benchmark designs perform better on the simulation dataset than they do on real data, while the accuracy of the DNNs-enabled methods on real data is higher than that on the simulation dataset. This is attributed to the fact that we only consider additive variations and errors in the simulation dataset while real data contains transformations like scaling and shifting, and the DNNs are relatively insensitive to transformations so that they can extract invariant features better on real data than they do on the simulation dataset.
\begin{table}
\centering
\caption{Accuracy for Different Methods on Real Data}
\begin{tabular}{c|c|c} 
\hline
Method&Accuracy (\%)&Time Complexity (s)\\
\hline
KNN&87.3&2.2\\
\hline
Logistic&86.8&23\\
\hline
SVM&87.8&129.5\\
\hline
CNN-r&97.4&10.9\\
\hline
RNN-r&95.5&12.2\\
\hline
CRNN-r&98.1&10.2\\
\hline
\end{tabular}\label{tableRealData}
\end{table}
\section{Conclusion}
This work studied CSI-based authentication algorithms that require no a priori information in a time-variant communication environment. The DNNs were introduced to build authenticators which connect CSI to its authenticated identity. To begin with, we built a CNN-enabled authenticator that can be invariant to transformations caused by environment changes and practical imperfections. Next, a RNN-enabled classifier was established to capture the spectral dependencies in CSI. In addition, we proposed to use the CRNN, which is a combination of the CNN and the RNN, to extract both the local and contextual messages in CSI for authentication. Moreover, we extended these DNNs-enabled authenticators into semi-supervised ones, wherein both the labeled and unlabeled data can be exploited to train the networks. According to the simulation and experimental results, all these DNNs-enabled authenticators have significant performance gains over the benchmark schemes, while the combination of the CNN and the RNN can not only enhance the authentication accuracy but also accelerate the convergence. Also, the DNNs-enabled approaches can perform better on real data than they do on the simulation dataset while the results are the opposite for the benchmark designs. This is because real data contains transformations while the simulation dataset does not, and the DNNs are relatively insensitive to transformations. In addition, the simulation results showed that the proposed semi-supervised DNNs-enabled methods can deliver excellent performance even with limited labeled channel observations.
\bibliographystyle{IEEEtran}
\bibliography{newbib}
\end{document}